\documentclass{article}
\PassOptionsToPackage{numbers}{natbib}

\usepackage[final]{neurips}
\usepackage{cite}
\usepackage{amsmath,amssymb,amsfonts}
\usepackage{graphicx}
\usepackage{subfigure}
\usepackage[parfill]{parskip}
\usepackage{multirow}
\usepackage{hhline}
\usepackage{xcolor}

\usepackage{algorithm}
\usepackage[hidelinks]{hyperref}
\usepackage{algpseudocode}
\usepackage{textcomp}
\usepackage{xcolor}
\usepackage{booktabs} 
\def\BibTeX{{\rm B\kern-.05em{\sc i\kern-.025em b}\kern-.08em
    T\kern-.1667em\lower.7ex\hbox{E}\kern-.125emX}}

\newcommand{\E}{\mathbb{E}}
\newcommand{\probP}{\mathbb{P}}

\newcommand{\Xx}{\textbf{$\mathcal{X}$}}
\newcommand{\Y}{\textbf{$\mathcal{Y}$}}

\newcommand{\W}{\mathcal{W}}

\newcommand{\Ss}{\mathcal{S}}

\newcommand{\D}{\textbf{\textit{D}}}

\DeclareMathOperator*{\argmin}{arg\,min}

\newtheorem{theorem}{Theorem}[section]

\newtheorem{definition}[theorem]{Definition}

\begin{document}

\title{Achieving Fairness Across Local and Global Models in Federated Learning}

\author{Disha Makhija \\
  The University of Texas at Austin
  \And 
  Xing Han \\
  John Hopkins University
  \And 
  Joydeep Ghosh \\
  The University of Texas at Austin
  \And 
  Yejin Kim\\
  The University of Texas Health at Houston}

\date{%
    $^1$University of Texas at Austin\\%
    $^2$John Hopkins University\\[2ex]%
    $^3$University of Texas Health at Houston\\[2ex]%
    \today
}

\maketitle

\begin{abstract}
Achieving fairness across diverse clients in Federated Learning (FL) remains a significant challenge due to the heterogeneity of the data and the inaccessibility of sensitive attributes from clients' private datasets. This study addresses this issue by introducing \texttt{EquiFL}, a novel approach designed to enhance both local and global fairness in federated learning environments. \texttt{EquiFL} incorporates a fairness term into the local optimization objective, effectively balancing local performance and fairness. The proposed coordination mechanism also prevents bias from propagating across clients during the collaboration phase. Through extensive experiments across multiple benchmarks, we demonstrate that \texttt{EquiFL} not only strikes a better balance between accuracy and fairness locally at each client but also achieves global fairness. The results also indicate that \texttt{EquiFL} ensures uniform performance distribution among clients, thus contributing to performance fairness. Furthermore, we showcase the benefits of \texttt{EquiFL} in a real-world distributed dataset from a healthcare application, specifically in predicting the effects of treatments on patients across various hospital locations.

\end{abstract}

\section{Introduction}\label{sec:intro}
Fairness in federated learning (FL) is an evolving area of research that seeks to ensure equitable outcomes for all participants in the FL process. Achieving fair FL involves ensuring that both local training, as well as the collaboration and aggregation of information across clients, maintain fairness. The distributed and heterogeneous nature of data sources in FL presents significant challenges, making it much harder to achieve fairness compared to centralized systems. 

Recent works on addressing fairness in FL have primarily focused on \textit{performance fairness}, which aims to achieve a more uniform accuracy distribution across clients. For instance, Ditto \citep{li2021ditto} proposes local training with regularization that encourages personalized models to approximate the optimal global model. Q-FFL \citep{li2019fair} minimizes an aggregate reweighted loss, assigning higher relative weights to devices with higher losses. PropFair \citep{zhang2022proportional} is designed to achieve proportional fairness in FL by balancing the average performances across all clients and ensuring satisfactory performance for even the worst-performing clients. Similarly, \citep{mohri2019agnostic} proposes agnostic FL, aiming to train models that do not overfit the data of any particular client at the expense of others. Another line of research focuses on \textit{collaboration fairness}, where each client's contribution is evaluated, and higher contributions receive higher rewards \citep{xu2020towards}. Contributions can be assessed using naive methods based on relative data volume and variety or more advanced techniques such as the Shapley value \citep{wang2020principled}, local credibility mutual evaluation \citep{lyu2020collaborative}, or level-wise measurement of contribution as seen in hierarchically fair FL \citep{zhang2020hierarchically}. 

Despite advancements in performance fairness and collaborative fairness, achieving \textit{model fairness} during FL procedures remains a significant and ongoing challenge. Model fairness refers to a model's ability to make non-biased predictions for any group or individual without discriminating against individuals' protected (sensitive) attributes such as race, gender, etc. However, training and evaluating models for fairness typically necessitate direct access to user-specific protected attributes. In the FL context, client data, including protected attributes, is strictly private and inaccessible outside the client environment. This confidentiality impedes the evaluation and assurance of fairness in the global model across all clients, creating an inherent conflict between fair model training and the decentralized nature of FL. Consequently, centralized methods for mitigating unfairness are hard to utilize in the FL setting.
Furthermore, variations in the distribution of protected attributes due to data heterogeneity across clients and insufficient representation within clients can lead to compromised fairness assessments based on local data alone. Even if local models exhibit fairness concerning protected attributes, this does not necessarily extend to fairness in the aggregated global model \citep{hamman2023demystifying}.  Finally, ensuring fairness in FL is even more complicated by the need to balance performance-fairness trade-offs across diverse client datasets, which often have varying data quality and distributions. This heterogeneity can result in models that perform well for some clients while disadvantaging others, exacerbating existing biases \citep{fairfed}.

With the above-mentioned challenges, we identify key research questions that have not yet been addressed by prior works within the community to ensure model fairness during collaborative training procedures. By proposing \texttt{EquiFL}, we take an important step in enhancing both local and global fairness in FL when client data contains protected attributes while simultaneously ensuring balanced performance across clients. We summarize our key contributions as follows:
\begin{itemize}
    \item \texttt{EquiFL} incorporates a fairness term into the local optimization objective for each client, aiming to explicitly balance local performance with fairness. This explicit bias mitigation tackles the challenges arising from heterogeneous data and varied distributions of protected attributes across clients. 
    \item \texttt{EquiFL} prevents the propagation of sensitive information during collaborative training and allows the personalization of prediction parts for each client.
    \item Experimental findings underscore the efficacy of \texttt{EquiFL} across various benchmarks. This is evidenced by experiments conducted on prominent fairness datasets, along with a case study focusing on equitable treatment outcomes based on clients' protected attributes for a healthcare application.
\end{itemize}
\section{Background}
We consider an FL scenario with $N$ clients, each possessing a distinct data distribution denoted by $\D_i, \forall i \in [1, ... N]$. The data available on each client comprises of a set of variables represented as $(\Xx^i, \Ss^i, \Y^i)$, where $\Ss^i$ refers to the protected or sensitive attribute under consideration, $\Xx^i$ denotes the other features to be utilized for prediction, and $\Y^i$ represents the observed outcome for specific instances on client $i$. Due to the diverse environments in which these clients operate, it is generally the case that the joint data distributions are non-IID, $\D_i \neq \D_j$ for any two clients $i$ and $j$. Moreover, if $p(\Ss^i)$ denotes the marginal distributions of the sensitive attribute on the client $i$, we also have non-IID marginal distributions $p(\Ss^i) \neq p(\Ss^j)$. In this context, we provide specific definitions of the concepts of \textit{local fairness} and \textit{global fairness} as follows.

\begin{definition}[Local Fairness]
\label{def:local_fairness}
Local fairness refers to the disparity exhibited by the model deployed on the client side when evaluated on that specific client's dataset.
\end{definition}

\begin{definition}[Global Fairness]
\label{def:global_fairness}
Global fairness refers to the disparity shown by the global model when evaluated on the dataset comprising data from all clients.
\end{definition}

Local fairness is crucial for each client because the client model is the one actively used in practice on that specific client. Therefore, it is important to assess the fairness and bias of the deployed model. On the other hand, global fairness is important as it indicates how the global model will perform on any new client. Global fairness has been the metric of primary focus in recent literature on fair FL~\citep{Du2020FairnessawareAF, Abay2020MitigatingBI, fairfed}.

\subsection{Federated Learning}
The standard FL procedure, FedAvg~\citep{fedavg_mcmahan}, iteratively trains a global model $\bar{f}$ parameterized by $\bar{\mathcal{W}}$ at the server. First, the procedure learns local client model parameters $\mathcal{W}_i$ for each client $i$ by optimizing the following local objective function,
\begin{equation}\label{eqn:fl}
    \W_i =  \argmin_{\W_i} \E_{(x_j,s_j,y_j) \sim \D_i }\ell(y_j, f_i(x_j; \mathcal{W}_i)).
\end{equation}
where $\ell(.)$ is any loss function. Subsequently, an element-wise average of all the client model parameters is computed to obtain the corresponding weights, $\bar{\W}$, of the aggregated model at the server. This aggregated model is then shared back with the local clients for further training. The entire procedure is repeated for $T$ communication rounds to learn a final global model, $\bar{f}(.,\bar{\W})$, to be used at all the clients for prediction.

\subsection{Fairness}
In machine learning, unfairness typically refers to a model discriminating against certain groups of people, such as those defined by race, age, gender etc. While our method can be used for any pre-specified notion of fairness, in this paper we evaluate the fairness of a machine learning model using the criterion known as disparate impact. It is important to note that a model cannot be fair under all fairness metrics simultaneously, as these definitions often conflict with one another~\citep{survey_in_processing, conflict_between_individual_group}. 

Disparate impact refers to a situation where the model disproportionately discriminates against certain groups, even if it doesn't explicitly use the sensitive attribute for predictions but relies on proxy attributes instead. While disparate impact can be measured in multiple ways, we consider two specific metrics known as demographic parity and equal opportunity which we define below for clarity. Our approach, though, can be readily extended for other outcome based fairness metrics in the literature.

\begin{definition}[Demographic Parity]
\label{def:demographic_parity}
Demographic parity is used to ensure that the outcome of a predictive model is independent of a specific protected attribute, i.e., the probability of a positive outcome (e.g., being approved for a loan) should be the same for all groups defined by the protected attribute. Mathematically, a model satisfies demographic parity if 
:
$$ \probP(f(x) = 1 | \Ss = s) = \probP(f(x)=1)$$
for all values of the protected attribute $s$ if $f(x)$ denotes the predicted outcome of $x$. The disparity in the demographic parity, denoted by $\Delta$-DP, ideally should be zero and is measured by :
\begin{equation}\label{eqn:dp}
 \max_{s, s' \in \Ss} | \probP(f(x) = 1 | \Ss = s) - \probP(f(x) = 1 | \Ss = s')|.
\end{equation}
\end{definition}

\begin{definition}[Equal Opportunity]
\label{def:equal_opportunity}
Equal opportunity ensures individuals in different demographic groups who qualify for a positive outcome (e.g., loan approval) have an equal chance of receiving that outcome. Specifically, it requires that the true positive rate (TPR) be the same across all groups. Mathematically, a model satisfies equal opportunity if :
$$ \probP(f(x) = 1 | Y= 1, \Ss = s) = \probP(f(x)=1|Y= 1)$$
for all values of the protected attribute $s$ if $Y$ denotes the actual outcome and $f(x)$ denotes the predicted outcome of $x$. The difference in equal opportunity denoted $\Delta$-EO, also has an ideal value of zero and is given by :
\begin{equation}\label{eqn:eo}
\max_{s, s' \in \Ss} | \probP(f(x) = 1 | Y =1,  \Ss = s) - \probP(f(x) = 1 | Y =1, \Ss = s')|.
\end{equation}
\end{definition}

\begin{figure*}[!htb]
    \centering
    \subfigure[]{\includegraphics[width=0.3\textwidth]{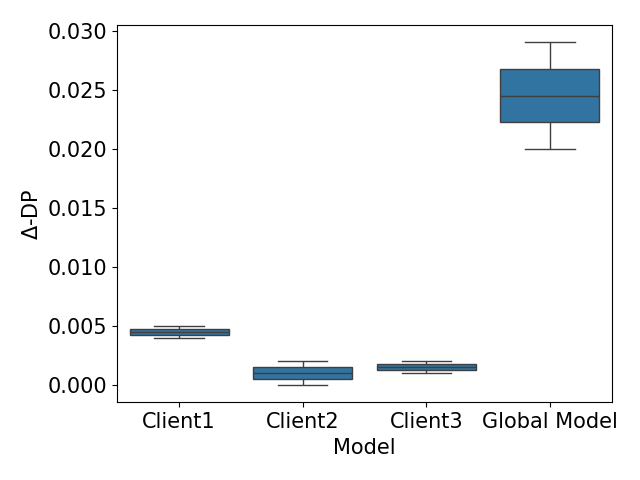}}
    \subfigure[]{\includegraphics[width=0.3\textwidth]{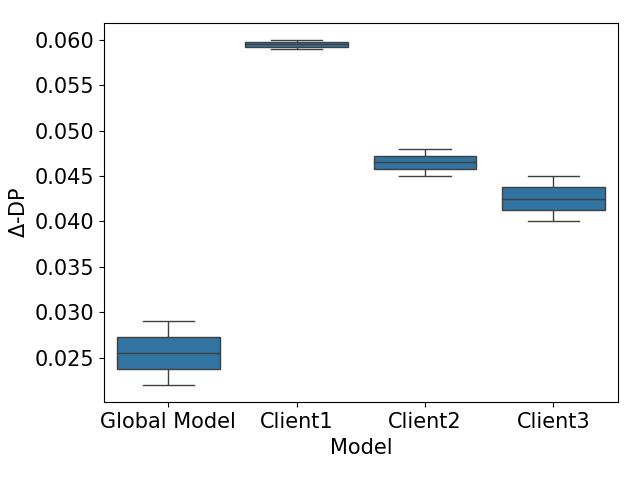}}
    \subfigure[]{\includegraphics[width=0.3\textwidth,height=1.75in]{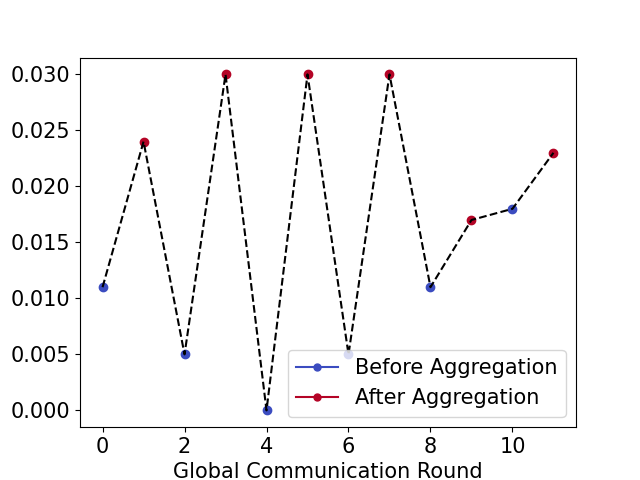}}
    \caption{Comparison of the $\Delta$-DP of the local and the global models for a 3 client setting on the Adult dataset in (a) and (b). Figure (a) shows imposing local fairness doesn't guarantee a fair global model, and figure (b) shows that imposing global fairness doesn't ensure fairness in local models. Figure (c) shows the change in the local fairness of a single client before and after the aggregation step in FL highlighting how the aggregation step might propagate bias.}
    \label{fig:local_global_fairness}
\end{figure*}

\subsection{Challenges}
Achieving fairness in an FL setting presents a unique set of challenges that stem from the fundamental nature of FL systems. Here, we delve into the primary obstacles that make fairness in FL both critical and complex.

\textbf{Local fairness does not imply global fairness.} Local fairness can be achieved through various methods, both explicit and implicit, such as regularization techniques \citep{wan2021modeling, olfat2020flexible, wang2021enhancing}, constrained optimization \citep{perrone2021fair, cotter2019optimization}, and learning disentangled representations \citep{sarhan2020fairness, park2020readme, oh2022learning}. These methods effectively address fairness within the local data distribution of each client. However, mitigating local unfairness typically focuses solely on the local context and does not ensure fairness at a global level across the entire federated learning system. Consequently, the global model may still exhibit unfairness when considering the aggregated data from all clients. Moreover, a model that appears fair when assessed locally on a particular client's data might be exacerbating bias globally. This can be caused, for example, by widely varying demographic proportions across clients. For example, a health institution operating in a predominantly Asian population city might have a locally fair model that disadvantages other racial groups on a larger scale. This is shown in Fig.~\ref{fig:local_global_fairness} (a) where local models for all clients are made fair by incorporating a fairness regularizer into the learning objective, yet the global model remains unfair.

\textbf{Global fairness does not imply local fairness.} On the other hand, learning a globally fair model from local models also does not guarantee fairness with respect to individual local data distributions. This can occur if a model that seems fair globally is actually making biased predictions for individual clients, which balance each other out when viewed globally. For example, consider a model used to predict patient treatment plans. The model might appear fair when considering all patient data across different hospitals, but it could still be biased against certain groups at specific hospitals. This could happen if the model favors younger patients at one hospital and older patients at another, creating an illusion of fairness at the global level while maintaining local biases. This is demonstrated in Fig.~\ref{fig:local_global_fairness} (b), where the global model is generated in each communication round by obtaining client model weights that maximize both overall performance and fairness. These phenomena and their information-theoretic explanations were studied in detail in \citep{hamman2023demystifying}.

\textbf{FL propagates bias.} In FL, algorithmic bias from one participant can spread to others, even if they don’t have biased data. This typically occurs because the biased participant unknowingly introduces bias into a few model parameters, which are then shared with everyone during the model merging process. The aggregated global model, containing these biased parameters, is sent back to all clients after merging. This cycle repeats over many rounds, causing the global model to increasingly rely on the biased parameters. Consequently, the FL model can become more biased than a model trained centrally on all combined data, even if most participants have unbiased data.
Figure~\ref{fig:local_global_fairness} (c) illustrates this phenomenon for a particular client, showing how the client’s local fairness, as measured by the $\Delta$-DP metric, worsens after receiving the aggregated model from the server. A similar phenomenon was also observed in~\citep{chang2023bias}.

\section{Methodology}
\begin{figure*}
\centering
    \centering
    \includegraphics[scale=0.4]{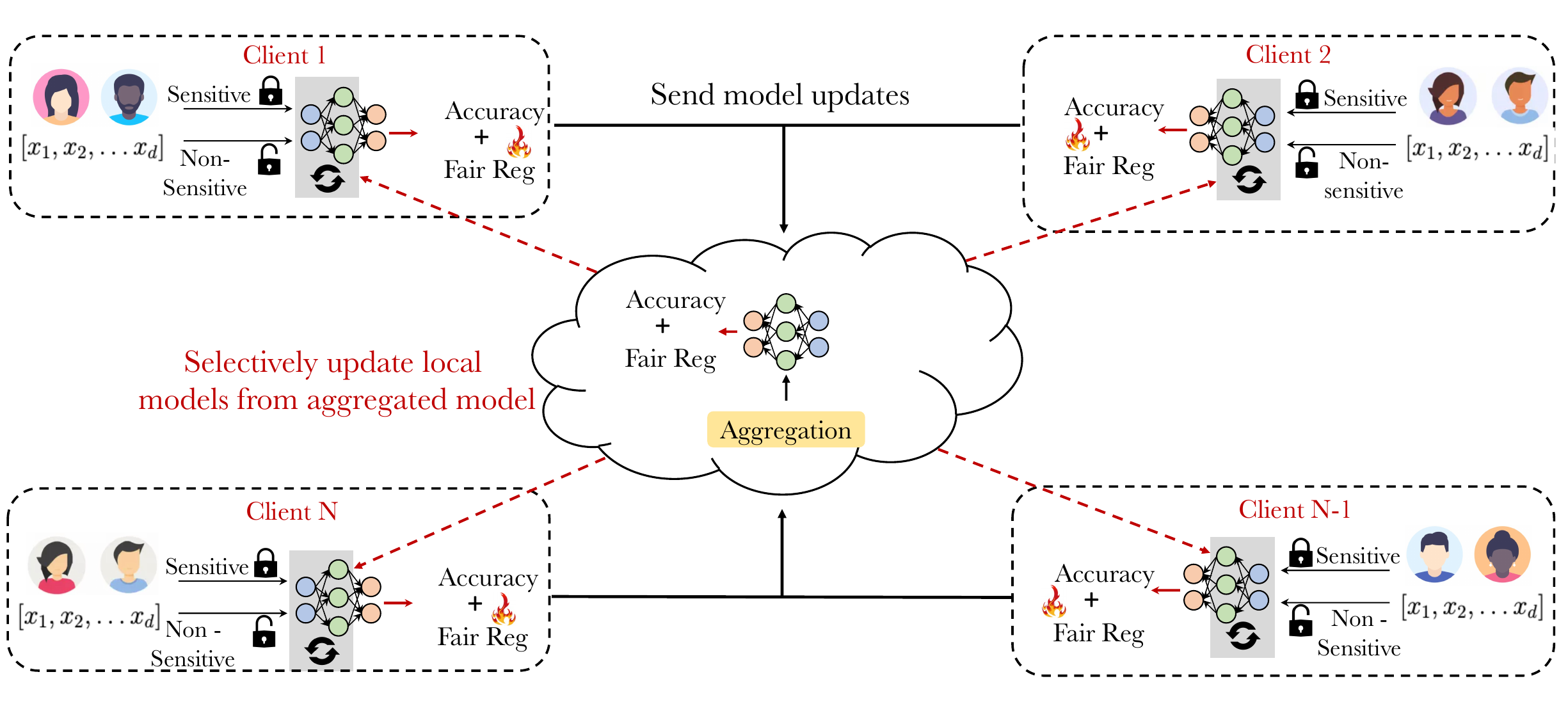}
    \caption{An overview of the proposed method showing $N$ clients, a server and the key steps in the procedure.}
    \label{fig:overview}
\end{figure*}

In this section, we first present the key insights that our learning algorithm is based on. Following that, we will provide a detailed specification of the algorithm, highlighting its components, mechanisms, and operational steps. An overview of the method is shown in Fig.~\ref{fig:overview}.

\subsection{Key Insights}
\textbf{Mitigating local unfairness is important.} After the FL training procedure, models are deployed locally, making it essential for these models to prevent unfair outcomes for the local client populations. However, the heterogeneous nature of data distributions and sensitive attributes across clients, combined with the lack of access to this information, complicates the task of ensuring performance-fairness trade-off through a global mechanism at the server. Specifically, the target population for the local client model, $f_i$, at the $i^{th}$ client is $\D_i$, however, existing methods typically build models based on performance and fairness evaluation using the combined distribution given by $\bar{\D} = \cup_{i \in |N|} \D_i$ which can prove to be detrimental to individual clients' models. Therefore, it's crucial for the local training process to explicitly address and mitigate unfairness at each client.

In our approach, we incorporate a fairness term into the local optimization objective for each client, prioritizing the balance between local performance and fairness during the local optimization step of the learning. In particular, instead of solving the optimization objective given by Equation~\ref{eqn:fl} locally at the $i^{th}$ client, we modify the objective function to incorporate a fairness term, as below :
\begin{multline}\label{fl_fairness}
    \W_i =  \argmin_{\W_i} \E_{(x_j,s_j,y_j) \sim \D_i }\ell(y_j, f_i(x_j; \mathcal{W}_i)) \\
    - \mu \text{Fairness}(f_i(.; \W_i)).    
\end{multline}
where Fairness$(f_i(.; \W_i))$ represents the fairness of model $f_i$ under the parameters $\W_i$, and $\mu$ is the weight corresponding to the importance of the fairness term in the objective. While any outcome based optimizable fairness metric relevant to the application can be used in the objective; we utilize the difference in demographic parity for the experiments. This explicit approach to mitigating bias at the client level is effective, requires minimal modifications to the objective function and does not increase the resource overhead for each client.

\textbf{Preventing the propagation of bias is critical.} In unfair models, bias travels from the input to the output through the model's weights. For example, in neural network-based models, the sensitive attribute affects the output via at least one path from the input to the output, where a path consists of connections between successive layers. Disrupting this flow of sensitive information to the output can disentangle the output from the sensitive information present in the input. While the local training procedure with fairness regularization discussed above, achieves fairness in the local models, the aggregation of model parameters in each communication round and the initialization of client models with the aggregated global model after each communication round can introduce bias from other clients' models into the local client. This process can cause the updated local models to become unfair on their specific local data distribution.

Consider each client's local model $f_i$ to be a $m$-layered neural network, then the function $f_i(x)$ can be written as -
\small
$$f_i(x) = W_m^i \psi_{m-1} (W_{m-1}^i \psi_{m-2} ( \dots \psi_1(W_1^i x + b_1^i)) + b_{m-1}^i ) + b_m^i $$
\normalsize
where $W_l^i$ and $b_l^i$ correspond to the weight matrix and the bias term of the $l^{th}$ layer of the network, with $\W_i = \{ W_1^i,b_1^i, \dots ,W_m^i, b_m^i \}$, the path from the input to the output is affected by the weight matrices as well as the bias terms. Even though the number of parameters in the bias terms is much smaller than in the weight matrices, the addition of the bias term to each layer and its propagation through the end makes it a significant contributor to the output of the model.

To prevent the unfairness incidentally acquired by the global model from propagating to the clients' local models, our method selectively updates the client model parameters from the aggregated model during each communication round. Specifically, the local models refrain from updating the parameters of the last layer (predictor), and for every other layer, the local models copy only the weight matrices from the global model, while keeping the bias parameters unchanged. Therefore, parameters related to the prediction layer and bias terms in each layer are optimized solely using the local optimization objective defined in Equation~\ref{fl_fairness} on local data, which also considers local fairness. This approach allows local models to collaboratively learn shared data representations across clients through weight matrices, while also retaining the flexibility to adjust bias terms. This adjustment helps cancel any bias that may have crept into the output of each layer of the model, ensuring fair representations at all layers and ultimately at the output. By selectively updating in this manner, local models can prevent the propagation of sensitive information through layers acquired during global model aggregation, thereby avoiding the learning of unfair patterns. This straightforward technique enables local models to leverage collaboration with other clients while maintaining necessary personalization to ensure fairness and performance aligned with local distribution characteristics.

\subsection{Algorithm}
The overall training procedure of the proposed method consists of $T$ communication rounds between the server and the clients. In each communication round, clients first perform local optimization, followed by collaboration through the server. The local optimization at each client involves solving the local optimization problem on the local data and updating all the local parameters $\W_i = \{ W_1^i,b_1^i, \dots ,W_m^i, b_m^i \}$ for $E$ epochs. Specifically, each client $i$, minimizes the objective function that is given by : 
\begin{align}\label{eqn:local_opt}
    \W_i^* &=  \argmin_{\W_i} \E_{(x_j,s_j,y_j) \sim \D_i }\ell(y_j, f_i(x_j; \mathcal{W}_i))  \nonumber \\
    &+ \mu \max_{ \substack{(s,s') \sim p(\Ss^i) \\ s \neq s'}} | \E[f_i(x) = 1 | \Ss = s] - \E[f_i(x) = 1 | \Ss = s']|. 
\end{align}
The optimized $\W_i^*$ from each client are uploaded to the server where an element-wise aggregation of all model parameters is performed to construct a global model $\bar{f}$ parameterized by $\bar{\W} = \{ \bar{W}_1,\bar{b}_1, \dots ,\bar{W}_m, \bar{b}_m \}$, which for round $(t)$ are obtained as follows:
\begin{equation}
    \bar{W_l}(t) = \sum_{i=1}^{N} \dfrac{n_i}{\sum_{i'=1}^{N} n_{i'}} W_l^i(t) ; \quad \bar{b_l}(t) = \sum_{i=1}^{N} \dfrac{n_i}{\sum_{i'=1}^{N} n_{i'}} b_l^i(t),
\end{equation}
for all layers of the neural network, $l \in [1, ..., m]$. The aggregated parameters are then sent to all clients for the next round of training. Clients initialize their local models from the global model and continue with local optimization. In our method, since the clients only copy the parameters corresponding to the weight matrices, at the beginning of local optimization in round $(t+1)$ at client $i$, we have
\begin{equation}\label{eqn:local_update}
    W_l^i(t+1) = \bar{W_l}(t), \quad \forall l \in [1, ..., m-1].
\end{equation}
The other parameters in the last layer ($m^{th}$) and the bias terms remain unaffected. The pseudo-code for this procedure is given in Algorithm~\ref{algo}.

\begin{algorithm}[!tb]
   \caption{\texttt{EquiFL} Algorithm}
   \label{algo}
   \begin{algorithmic}
   \State \textbf{Input:} number of clients $N$, number of global communication rounds $T$, 
   number of local epochs $E$, parameter $\mu$. 
   \State \textbf{Output:} Final global model $\bar{f}(.,\bar{\W}(T))$ and local models ${f_i}(.,{\W_i}(T))$
    \\\hrulefill
   \State {\bfseries At Server - }
   \State Initialize $\bar{\W}(0)$
   \For{$t=0$ {\bfseries to} $T-1$}
    \State Select a subset of clients $\mathcal{N}_t$ 
    \For{each selected client $i \in \mathcal{N}_t$}
    \State $\W_i(t+1) =$ \textbf{LocalTraining}$( \bar{\W}(t), \mu)$
    \EndFor
    \State $\bar{\W}(t+1) = \dfrac{1}{\sum_{j \in \mathcal{N}_t} n_j} \sum_{i \in \mathcal{N}_t} n_i \W_i(t+1)$
   \EndFor
   \State Return $\bar{\W}(T), {\W_1}(T) \dots {\W_N}(T)$
   \\\hrulefill
    \State \textbf{LocalTraining}$( \bar{\W}(t), \mu)$
    \State Initialize $\W_i(t+1)$ using $\bar{\W}(t)$ according to Equation~\eqref{eqn:local_update}
    \For{each local epoch}
    \State Update $\W_i(t+1)$ by solving objective in ~\eqref{eqn:local_opt}
    \EndFor
    \State Return $\mathcal{W}_i(t+1)$ to the server 
\end{algorithmic}
\end{algorithm}
\section{Experiments}
In this section, we present a comprehensive experimental evaluation of our proposed method, \texttt{EquiFL}, and compare its performance with several baseline approaches. We begin by detailing the experimental setup, including datasets, data partitioning, evaluation metrics, and baseline methods, and then present the results of our experiments highlighting the effectiveness of \texttt{EquiFL}.

\subsection{Experimental Setting}
\textbf{Datasets} We consider three widely used binary classification datasets that are well-known in the fairness literature for evaluating and benchmarking our method and the baselines. Adult dataset (ACSIncome Dataset)~\citep{adult_dataset} is based on 1994 U.S. census data and contains information about approximately 30,000 individuals. The task is to predict whether an individual earns more than \$50,000 per year, with the sensitive attribute being the sex of the individual. The COMPAS dataset~\citep{compas_dataset} is a recidivism risk assessment tool developed by Northpointe, used by judges to inform sentencing decisions. It includes information about individuals, with the task being to predict whether an individual will re-offend. The sensitive attribute in this dataset is race. And lastly, Heritage Health dataset~\footnote{https://www.kaggle.com/c/hhp} comprises data on around 51,000 patients. The task is to predict the Charleson Index, an indicator of a patient's 10-year survival rate. For this dataset, we consider age and gender as the protected attributes in two different experiments.

\textbf{Baselines} We consider the following state-of-the-art fair federated learning methods as baselines: i) FedAvg, the conventional federated learning procedure~\citep{fedavg_mcmahan}; ii) FairFed, an FL procedure that adjusts the aggregation weights for local client models to create a fairer global model~\citep{fairfed}; iii) LFT+FedAvg, which uses a local reweighing approach to develop locally fair solutions~\citep{bhaskaruni2019improving}; and iv) FedFB, which uses a local reweighting mechanism for groups to create fair models~\citep{fedfb}.

\textbf{FL Simulation} To simulate an FL environment, non-IID partitions of the dataset equal to the number of clients in the simulation are created by sampling a fraction of instances $p_{v,i} \sim Dir(\alpha)$ to allocate to the $i^{th}$ client for each value $v$ of each sensitive attribute. Here, $\alpha$ controls the degree of data heterogeneity across the clients, and to achieve a realistic setting, we set the $\alpha$ values differently for each client. For the experiments shown in this section, we partition the data into 5 clients and assign $\alpha$ values of $[0.1, 0.2, 1, 10, 0.5]$. This creates data partitions across clients with varied proportions of the sensitive attribute values. In the case of a binary sensitive attribute like gender, $\alpha$ values result in a distribution where the proportion of one value (say male) is $[0.99, 0.95, 0.65, 0.5, 0.90]$. This approach provides a more realistic setting than using the same $\alpha$ value for each client.  

\textbf{Training protocol} After partitioning the datasets into clients' local datasets, each dataset is further divided into training, validation, and test splits with corresponding ratios of 70:15:15. All experiments are conducted over 5 rounds, with performance metrics reported on a held-out test dataset. The parameter $\mu$ is set to 1 for all runs. Hyperparameters such as learning rate, batch size, and the number of local epochs are selected from the ranges $[1e-4, 1e-3, 1e-2, 1e-1]$, $[256, 512, 1024, 2048]$, and $[5, 10, 20]$, respectively, by tuning over the validation set for all methods. The entire procedure is configured to run for 100 communication rounds using the Adam optimizer. All models are trained on a machine with 4 GeForce RTX 3090 GPUs, each with 24GB of memory.

\subsection{Results} The results for the performance comparison between \texttt{EquiFL} and the baselines are included in Table~\ref{tab:results1} and Table~\ref{tab:results2} for clarity. The local metrics are obtained by averaging the local performance and local fairness over all clients, and the global metrics are obtained by calculating the same metrics from the global model on a global dataset. We observe that EquiFL outperforms the baselines in terms of the performance-fairness trade-off that it achieves. Specifically, we observe that the FedAvg algorithm, which does not explicitly aim for fairness, achieves higher test accuracy compared to methods that are specifically designed to enhance fairness. However, this increased accuracy comes at the expense of fairness, with the FedAvg algorithm mostly exhibiting the lowest fairness metric among the evaluated methods. Interestingly, \textit{our method generally achieves accuracy levels comparable to those of the FedAvg algorithm while maintaining significantly higher fairness}. This demonstrates that, for a given level of fairness comparable to the baseline methods, our approach can deliver superior performance.

\begin{figure}
\centering
    \centering
    \includegraphics[scale=0.4]{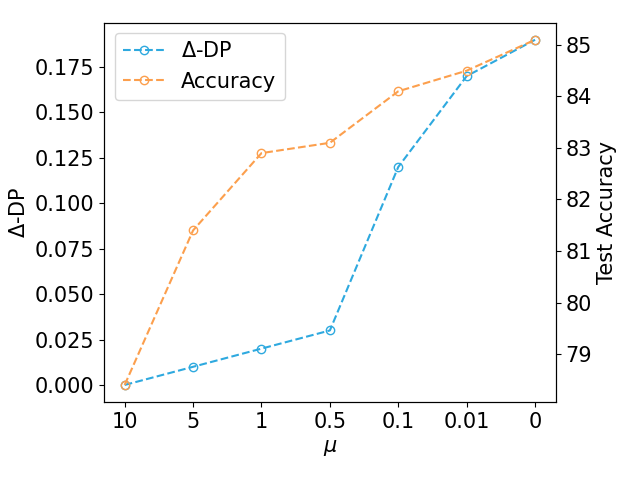}
    \caption{Change in performance and fairness with varying $\mu$ - the weight of the fairness term.}
    \label{fig:vary_mu}
\end{figure}

Since our method includes a fairness weight parameter $\mu$ in the local optimization procedure to address the fairness constraint, we analyzed how varying $\mu$ affects both accuracy and fairness. We conducted an ablation study using the Adult dataset with 10 clients, adjusting $\mu$ between 10 and 0, and recorded the resulting accuracy and $\Delta$-DP (fairness metric) for the local client models. The results, displayed in Fig.~\ref{fig:vary_mu}, show that increasing $\mu$ enhances fairness but decreases accuracy. The lowest accuracy, $78.8\%$ was observed when $\Delta$-DP $=0$ at $\mu = 10$. As $\mu$ decreased, accuracy improved. However, beyond $\mu = 1$ the gains in accuracy did not sufficiently compensate for the increase in $\Delta$-DP. Therefore, we selected $\mu = 1$ for all our experiments to balance accuracy and fairness effectively.

Furthermore, since the experimental results in Table~\ref{tab:results1} and Table~\ref{tab:results2} present findings for a 5-client FL setting, we extend our evaluation to demonstrate the performance of our method with an increased number of clients: 10, 20, 50, and 100, as shown in Table~\ref{tab:increasing_clients}. The reported results are the average local performance metrics on the Adult dataset. These results indicate that the average performance remains consistent even as the number of clients increases. This consistency is significant because, as the number of clients grows, the data points per client decrease, and maintaining performance under these conditions highlights the robustness of our method.

\begin{table}[!ht]
    \begin{center}
    \caption{Performance comparison (test accuracy and $\Delta$-DP) on the Adult and COMPAS dataset in both local and global models. The sensitive attribute is denoted next to the name of the dataset.}
    \label{tab:results1}
    \begin{small}
    \hspace*{-2.0cm}
    \addtolength{\tabcolsep}{-0.4em}
    \begin{tabular}{@{}lcccccccc@{}}
        \toprule
            \multirow{4}{*}[-0.5ex]{Method} &  \multicolumn{4}{c}{Adult (Gender)} & \multicolumn{4}{c}{COMPAS (Race)} \\[0.5ex]
            \cmidrule(lr){2-5} \cmidrule(lr){6-9} 
            & \multicolumn{2}{c}{Local Performance} & \multicolumn{2}{c}{Global Performance} & \multicolumn{2}{c}{Local Performance} & \multicolumn{2}{c}{Global Performance} \\[0.5ex]
            \cmidrule(lr){2-3} \cmidrule(lr){4-5} \cmidrule(lr){6-7} \cmidrule(lr){8-9} 
            & Accuracy ($\uparrow$) & $\Delta$-DP ($\downarrow$) & Accuracy ($\uparrow$) & $\Delta$-DP ($\downarrow$) & Accuracy ($\uparrow$) & $\Delta$-DP ($\downarrow$) & Accuracy ($\uparrow$) & $\Delta$-DP ($\downarrow$) \\[0.5ex]
            \midrule
        FedAvg & \textbf{83.9$\pm$1.5} & 0.07$\pm$0.02 & \textbf{83.7$\pm$0.3} & 0.10$\pm$0.02 & \textbf{70.8$\pm$1.0} & 0.16$\pm$0.01 & \textbf{69.9$\pm$0.7} & 0.13$\pm$0.02 \\[0.5ex]  
        FairFed & 83.0$\pm$0.5 & 0.052$\pm$0.01 & 82.5$\pm$0.1 & 0.08$\pm$0.01 & 69.3$\pm$0.2 & 0.15$\pm$0.04 & 68.7$\pm$0.2 & 0.14$\pm$0.01 \\[0.5ex]
        LFT + FedAvg & 80.4$\pm$0.05 & 0.06$\pm$0.02 & 81.3$\pm$0.2 & 0.06$\pm$0.02 & 60.9$\pm$0.07 & 0.11$\pm$0.01 & 60.4$\pm$0.1 & 0.11$\pm$0.04 \\[0.5ex]
        FedFB & 80.2$\pm$0.03 & 0.03$\pm$0.015 & 79.3$\pm$0.1 & 0.09$\pm$0.008 & 67.4$\pm$0.01 & 0.13$\pm$0.02 & 65.7$\pm$0.3 & 0.11$\pm$0.2 \\[0.5ex]
        EquiFL (Ours) & \textbf{83.8$\pm$0.6} & \textbf{0.03$\pm$0.008} & 82.2$\pm$0.7 & \textbf{0.02$\pm$0.004} & 69.5$\pm$0.05 & \textbf{0.115$\pm$0.01} & 69.3$\pm$0.08 & \textbf{0.09$\pm$0.01} \\[0.5ex]
        \bottomrule
    \end{tabular}
    \end{small}
    \end{center}
\end{table}

\begin{table}[!ht]
    \begin{center}
    \caption{Performance comparison (test accuracy and $\Delta$-DP) on the Heritage Health dataset in both local and global models. The sensitive attribute is denoted next to the name of the dataset.}
    \label{tab:results2}
    \begin{small}
    \hspace*{-2.0cm}
    \addtolength{\tabcolsep}{-0.4em}
    \begin{tabular}{@{}lcccccccc@{}}
        \toprule
            \multirow{4}{*}[-0.5ex]{Method} &  \multicolumn{4}{c}{Heritage Health (Gender)} & \multicolumn{4}{c}{Heritage Health (Age)} \\[0.5ex]
            \cmidrule(lr){2-5} \cmidrule(lr){6-9} 
            & \multicolumn{2}{c}{Local Performance} & \multicolumn{2}{c}{Global Performance} & \multicolumn{2}{c}{Local Performance} & \multicolumn{2}{c}{Global Performance} \\[0.5ex]
            \cmidrule(lr){2-3} \cmidrule(lr){4-5} \cmidrule(lr){6-7} \cmidrule(lr){8-9} 
            & Accuracy ($\uparrow$) & $\Delta$-DP ($\downarrow$) & Accuracy ($\uparrow$) & $\Delta$-DP ($\downarrow$) & Accuracy ($\uparrow$) & $\Delta$-DP ($\downarrow$) & Accuracy ($\uparrow$) & $\Delta$-DP ($\downarrow$) \\[0.5ex]
            \midrule
        FedAvg & 79.1$\pm$0.4 & 0.04$\pm$0.00 & 79.9$\pm$0.01 & 0.03$\pm$0.00 & \textbf{79.5$\pm$0.34} & 0.45$\pm$0.05 & \textbf{79.7$\pm$0.04} & 0.51$\pm$0.02 \\[0.5ex]
        FairFed & 80.4$\pm$0.2 & 0.035$\pm$0.01 & 80.3$\pm$0.1 & 0.02$\pm$0.01 & 79.5$\pm$0.34 & 0.45$\pm$0.05 & 79.7$\pm$0.04 & 0.51$\pm$0.02 \\[0.5ex]
        LFT + FedAvg & 78.7$\pm$0.4 & 0.04$\pm$0.01 & 78.0$\pm$0.6 & 0.06$\pm$0.01 & 76.4$\pm$0.8 & 0.40$\pm$0.04 & 76.1$\pm$1.0 & 0.47$\pm$0.02 \\[0.5ex]
        FedFB & 79.1$\pm$1.3 & 0.04$\pm$0.01 & 77.5$\pm$1.1 & 0.042$\pm$0.08 & 78.6$\pm$0.26 & 0.38$\pm$0.09 & 78.5$\pm$0.8 & 0.46$\pm$0.07  \\[0.5ex]
        EquiFL (Ours) & \textbf{80.5$\pm$0.2} & \textbf{0.03$\pm$0.0} & \textbf{80.7$\pm$0.1} & \textbf{0.02$\pm$0.00} & 79.0$\pm$1.7 & \textbf{0.33$\pm$0.07} & 78.9$\pm$0.16 & \textbf{0.38$\pm$0.01} \\[0.5ex]
        \bottomrule
    \end{tabular}
    \end{small}
    \end{center}
\end{table}

\begin{table}[!ht]
    \begin{center}
    \caption{Variation in performance with increasing number of clients.}
    \label{tab:increasing_clients}
    \begin{small}
    \hspace*{-0.8cm}
    \addtolength{\tabcolsep}{-0.4em}
    \begin{tabular}{@{}|l|ccc|@{}}
        \toprule
           \# clients &  Accuracy & $\Delta$-DP  &  $\Delta$-EO \\[0.5ex]
         \midrule
            10 & 83.0$\pm$0.1 & 0.02$\pm$0.01 & 0.03$\pm$0.007 \\[0.5ex]
            20 & 83.3$\pm$0.5 & 0.03$\pm$0.004 & 0.035$\pm$0.008 \\[0.5ex]
            50 & 82.5$\pm$0.5 & 0.03$\pm$0.01 & 0.037$\pm$0.01 \\[0.5ex]
            100 & 82.9$\pm$0.8 & 0.04$\pm$0.005 & 0.038$\pm$0.01 \\[0.5ex]
         \bottomrule
    \end{tabular}
    \end{small}
    \end{center}
\end{table}

\begin{figure}
\centering
    \centering
    \includegraphics[scale=0.4]{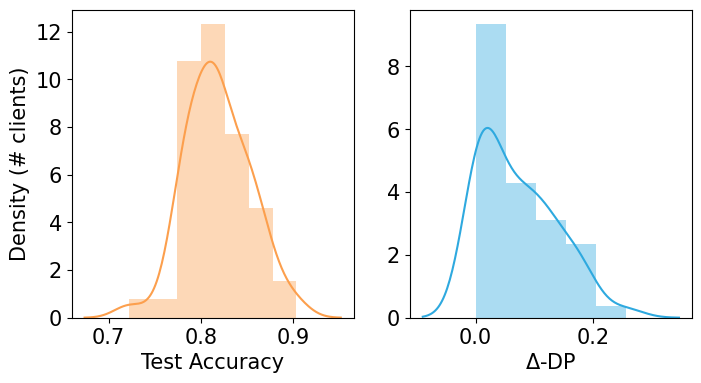}
    \caption{Accuracy and $\Delta$-DP distributions across 100 clients trained using EquiFL on Adult dataset.}
    \label{fig:perf_fairness}
\end{figure}

\textbf{Performance Fairness.} Due to the heterogeneous nature of the data across different clients,  minimizing an aggregate loss in a large FL network with many clients, can disproportionately advantage or disadvantage the model performance on some clients. For instance, although the overall accuracy may be high on average, there is no guarantee of accuracy for individual clients in the network, leading to significant variability in model performance. While our method does not explicitly aim for performance fairness, which involves achieving uniform performance across all participating clients, we observe that the distribution of both the test accuracy and fairness metrics obtained under our method is concentrated. This observation suggests that our approach inherently promotes balanced performance across different clients, even though it does not specifically target this outcome. As illustrated in Fig.~\ref{fig:perf_fairness}, the distribution of test accuracy and fairness metrics exhibits low variation. This indicates that our method effectively maintains a consistent level of fairness and performance across the federated learning network. The metrics' distribution concentration demonstrates that our approach can inherently provide equitable outcomes across various clients. This is particularly important because, in many critical applications, ensuring uniform performance and fairness across all clients is essential. Applications in healthcare, finance, and other sensitive fields require models that not only perform well on average but also maintain reliable and unbiased outcomes for all participating entities. The ability of our method to achieve this balance without explicitly targeting performance fairness underscores its robustness and suitability for real-world scenarios where data distribution and client needs can vary significantly.

\section{Case Study - Treatment Effect Estimation in Healthcare}
The experiments section demonstrates the effectiveness of our method in achieving a better trade-off between fairness and performance. To further validate the effectiveness of \texttt{EquiFL}, in real-world FL scenarios, particularly in critical applications where fairness is essential, we present a case study. This case study involves using our method to predict the effects of treatments on patient data obtained from various clinical trials conducted at different hospital locations.

\begin{figure}
\centering
    \centering
    \includegraphics[scale=0.4]{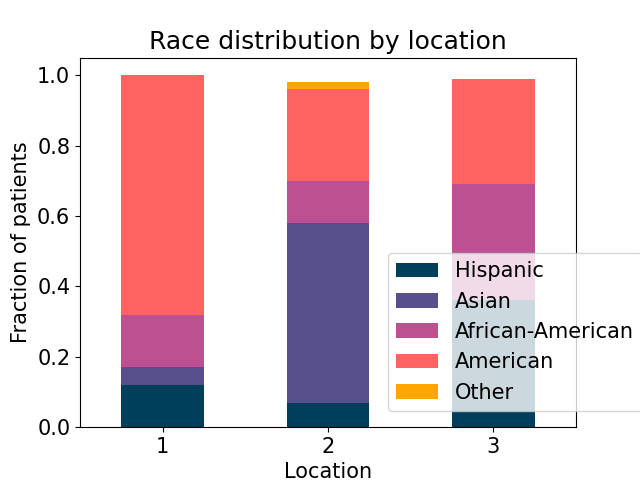}
    \caption{Distribution of patients by race across 3 locations in the ICH dataset.}
    \label{fig:race_dist}
\end{figure}

\textbf{Problem Setting and Dataset Description :} Clinical trials are research studies performed on human participants to evaluate the effectiveness and safety of medical treatments, such as drugs, therapeutic interventions, etc. The treatment effect estimation problem involves determining the impact of a specific treatment on patient outcomes. Accurate treatment effect estimation is crucial for developing effective and safe medical treatments. In this study, we consider the clinical trials used to develop the therapy for intracerebral hemorrhage (ICH)~\citep{ich_dataset}. Three different treatment protocols being administered at different hospital locations are considered : ATACH2 (Study:NCT01176565), MISTIE3 (Study:NCT01827046) and ERICH (Study:NCT01202864). Each hospital contributes patient-level pre-treatment measurements as features for prediction, with binary outcomes indicating the treatment's efficacy for each patient. Federated learning proves invaluable in learning collaboratively across these locations, especially considering the limited clinical trial data available locally at each hospital~\citep{makhija2024federated}. However, the natural heterogeneity in distributed data sources leads to significant variations in population demographics across different locations. These variations encompass factors like age, ethnicity, and gender among patient groups at different hospitals. Given the critical nature of the problem, it's important to develop high-performing and fair models capable of addressing these demographic differences across all locations.
\textbf{Data Distribution :} The dataset consists of $\sim$3200 patients distributed across 3 locations with each patient having 47 features. We consider each location as a client participating in the FL procedure. Because of the demographic distribution across various locations, clients possess varying data sizes and distributions of sensitive attributes. The distribution of the patients is included in Table~\ref{tab:dataset_dist}, and their characterization based on race is shown in Fig.~\ref{fig:race_dist}.

\begin{table}[!ht]
    \begin{center}
    \caption{Distribution of the ICH dataset.}
    \label{tab:dataset_dist}
    \begin{small}
    \hspace*{-0.8cm}
    \addtolength{\tabcolsep}{-0.4em}
    \begin{tabular}{@{}|l|c|c|c|@{}}
        \toprule
            &  Location 1 & Location 2  &  Location 3 \\[0.5ex]
         \midrule
          \# patients & 560 & 1000 & 1773 \\[0.5ex]
         \bottomrule
    \end{tabular}
    \end{small}
    \end{center}
\end{table}

\textbf{Results :} The experimental results, displaying both accuracy and fairness performance for local and global models, are presented in Table~\ref{tab:ich_results}. We consider two settings for this experiment: one where the sensitive attribute is gender and another where it is race. Our observations indicate that our method achieves a better accuracy and fairness trade-off in both settings. However, the performance difference is significantly more pronounced when the sensitive attribute is race. This is because race is more heterogeneously distributed across locations, as shown in Fig.~\ref{fig:race_dist}, and it has five distinct values so the maximum difference is much larger.

\begin{table*}[!ht]
    \begin{center}
    \caption{Predicting treatment effects of medical interventions on the ICH dataset with Gender and Race as sensitive attributes.}
    \label{tab:ich_results}
    \begin{small}
    \hspace*{-1.8cm}
    \addtolength{\tabcolsep}{-0.5em}
    \begin{tabular}{@{}lcccccccc@{}}
        \toprule
            \multirow{4}{*}[-0.5ex]{Method} &  \multicolumn{4}{c}{Sensitive Attribute - Gender} & \multicolumn{4}{c}{Sensitive Attribute - Race} \\[0.5ex]
            \cmidrule(lr){2-5} \cmidrule(lr){6-9} 
            & \multicolumn{2}{c}{Local Performance} & \multicolumn{2}{c}{Global Performance} & \multicolumn{2}{c}{Local Performance} & \multicolumn{2}{c}{Global Performance} \\[0.5ex]
            \cmidrule(lr){2-3} \cmidrule(lr){4-5} \cmidrule(lr){6-7} \cmidrule(lr){8-9} 
            & Accuracy ($\uparrow$) & $\Delta$-DP ($\downarrow$) & Accuracy ($\uparrow$) & $\Delta$-DP ($\downarrow$) & Accuracy ($\uparrow$) & $\Delta$-DP ($\downarrow$) & Accuracy ($\uparrow$) & $\Delta$-DP ($\downarrow$) \\[0.5ex]
            \midrule
        FedAvg & 92.9$\pm$0.03 & 0.01$\pm$0.001 & 88.9$\pm$0.02 & 0.023$\pm$0.01 & 91.3$\pm$0.74 & 0.02$\pm$0.001 & 86.2$\pm$1.7 & 0.17$\pm$0.1\\[0.5ex]  
        FairFed & 92.5$\pm$0.006 & \textbf{0.002$\pm$0.001} & 88.5$\pm$0.03 & 0.008$\pm$0.001 & 91.9$\pm$0.02 & 0.01$\pm$0.001 & 86.7$\pm$0.32 & 0.15$\pm$0.05 \\[0.5ex]
        EquiFL (Ours) & \textbf{93.2$\pm$0.02} & 0.007$\pm$0.001 & \textbf{89.1$\pm$0.05} & \textbf{0.000$\pm$0.001} & \textbf{92.2$\pm$0.2}& \textbf{0.01$\pm$0.001} & \textbf{89.0$\pm$0.10}& \textbf{0.05$\pm$0.008}\\[0.5ex]
         \bottomrule
    \end{tabular}
    \end{small}
    \end{center}
\end{table*}
\section{Related Works}
\textbf{Fairness in Centralized Setting} ~Model fairness is a critical concern in machine learning due to the ethical implications of biased decision-making systems. It is important given biased models towards any subgroups can perpetuate societal inequities, particularly in sensitive domains of healthcare \citep{obermeyer2019dissecting, hardt2016equality}, e-commerce \citep{burke2017multisided, ekstrand2018exploring}, finance \citep{hurley2016credit, chouldechova2017fair}, and technology \citep{bolukbasi2016man, raji2019actionable}. Various strategies exist to produce fair models in centralized setting, including pre-processing techniques to remove biases from data \citep{feldman2015certifying, abusitta2019multi}, in-processing methods that incorporate fairness constraints during model training \citep{kamishima2012fairness, zafar2017fairness, berk2017convex}, and post-processing approaches to adjust model outputs for fairness \citep{corbett2017algorithmic, dwork2018fairness, menon2018cost}. Additionally, adversarial debiasing and ensemble learning have been developed to enhance model fairness without sacrificing accuracy \citep{kenfack2023learning, bhaskaruni2019improving}.

\textbf{Fairness in FL} ~Different formulations of fairness have been studied in the FL setting, including performance fairness \citep{li2021ditto, li2019fair, zhang2022proportional} and collaboration fairness \citep{xu2020towards, wang2020principled, lyu2020collaborative}. In \texttt{EquiFL}, we address the problem of \textit{group fairness} which requires the model to perform comparably across groups defined by sensitive attributes, such as race, gender, or age \citep{ignatiev2020towards}. Recent studies have made significant strides in achieving group fairness within FL. A common research approach involves solving an optimization problem with fairness constraints in a distributed manner. Specifically, \citep{zhang2020fairfl} introduces a framework that uses a multi-agent reinforcement learning model and a secure aggregation protocol to achieve fairness and accuracy across demographic groups in FL. \citep{du2021fairness} proposes a framework that integrates kernel reweighing functions into both loss functions and fairness constraints to ensure high accuracy and fairness under unknown testing data distributions. \citep{galvez2021enforcing} introduces an algorithm that adapts the modified method of multipliers to enforce group fairness in private FL. This type of approach necessitates that each client shares statistics related to sensitive attributes from their local datasets with the central server. Moreover, \citep{abay2020mitigating} explored the efficacy of employing a global reweighting mechanism to enhance fairness. \citep{zeng2021improving} proposed an adaptation of the FairBatch debiasing algorithm \citep{roh2020fairbatch} for FL, where clients apply FairBatch locally, and weights are updated centrally each round. \citep{papadaki2021federating} introduced an algorithm to achieve mini-max fairness in FL. More recently, \citep{chang2023bias} demonstrates that FL can inadvertently propagate biases from a few parties against under-represented groups throughout the network, leading to fairness issues compared to standalone training on local data. \citep{fairfed} enhances group fairness by adjusting model aggregation weights based on local and global fairness measurements, demonstrating fairness improvements under heterogeneous data distributions. Compared with prior works, \texttt{EquiFL} effectively prevents bias propagation, resulting in enhanced local and global fairness. It also achieves balanced fairness and performance for each local client. Importantly, \texttt{EquiFL} maintains user privacy by not sharing statistics or model performance on subgroups divided by sensitive attributes.
\section{Conclusion}
In this work, we introduce a novel federated learning framework designed to enhance local fairness across diverse client datasets while maintaining global fairness. Our approach combines local fair model training with an effective collaboration mechanism to address disparities in performance and fairness caused by variations in data distributions among clients. The experimental results demonstrate that our method outperforms existing state-of-the-art fair federated learning techniques in terms of both accuracy and fairness metrics, successfully maintaining a balance between accuracy, local fairness, and global fairness. This indicates the potential of our method to be applied in real-world scenarios where equitable outcomes are critical, as also demonstrated in the case study on a real-world healthcare dataset. While our framework provides a substantial step forward in fair federated learning, several avenues for future research remain open. One promising direction is to incorporate differential privacy mechanisms to further protect the data privacy of clients while ensuring fairness. Secondly, real-world deployments and longitudinal studies are necessary to evaluate the robustness and scalability of the proposed framework in dynamic environments where client participation may vary over time. These future research directions will not only advance the field of federated learning but also contribute to the development of more equitable AI systems.

\bibliographystyle{plainnat}
\bibliography{refs}

\end{document}